%% file: acl_industry-2023.tex
\pdfoutput=1

\documentclass[11pt]{article}
\usepackage{graphicx}
\usepackage{pgf-pie} 
\usepackage{tabularx}

\usepackage[]{ACL_INDUSTRY-2023}
\usepackage{tabularx}

\usepackage{times}
\usepackage{latexsym}

\usepackage{amsmath}
\DeclareMathOperator*{\argmax}{argmax} 
\usepackage[T1]{fontenc}

\usepackage[utf8]{inputenc}

\usepackage{microtype}

\usepackage{inconsolata}

\usepackage{amsmath,array,graphicx}
\usepackage{kantlipsum}
\title{"We care": Improving Code Mixed Speech Emotion Recognition in Customer-Care Conversations}

\author{N V S Abhishek \\
  Department of Computer Science\\
  and Engineering \\
  IIT Bombay \\
  \texttt{nvsabhishek.india@gmail.com} \\
  \And
  Pushpak Bhattacharyya \\
  Department of Computer Science\\
  and Engineering \\
  IIT Bombay \\
  \texttt{pushpakbh@gmail.com} \\}

\begin{document}
\maketitle
\begin{abstract}
 Speech Emotion Recognition (SER) is the task of identifying the emotion expressed in a spoken utterance. Emotion recognition is essential in building robust conversational agents in domains such as law, healthcare, education, and customer support. Most of the studies published on SER use datasets created by employing professional actors in a noise-free environment. In natural settings such as a customer care conversation, the audio is often noisy with speakers regularly switching between different languages as they see fit. We have worked in collaboration with a leading unicorn in the Conversational AI sector to develop Natural Speech Emotion Dataset (NSED). NSED is a natural code-mixed speech emotion dataset where each utterance in a conversation is annotated with emotion, sentiment, valence, arousal, and dominance (VAD) values. In this paper, we show that by incorporating word-level VAD value we improve on the task of SER by 2\%, for negative emotions, over the baseline value for NSED. High accuracy for negative emotion recognition is essential because customers expressing negative opinions/views need to be pacified with urgency, lest complaints and dissatisfaction snowball and get out of hand. Escalation of negative opinions speedily is crucial for business interests. Our study then can be utilized to develop conversational agents 
 which are more polite and empathetic in such situations. 
\end{abstract}

\input{sections/intro.tex}

\input{sections/related.tex}

\input{sections/modeling.tex}

\input{sections/experiment.tex}

\input{sections/results.tex}

\section{Conclusion and Future Work}
In this paper, we discussed the effect of incorporating word-level VAD values on SER for the Natural Speech Emotion Dataset (NSED). We also described the steps involved in creating NSED. Frequent code-mixing and noisy environments are some the biggest challenges for performing SER on natural datasets. By incorporating word-level VAD values we were able to achieve an improvement of $2\%$ over the baseline in SER for negative emotions. In future, we look forward to expanding this dataset so that all the emotions have substantial examples. Our SER system can be used to develop conversational agents which generate polite and empathetic statements to pacify a frustrated/angry customer. Different unsupervised techniques for SER can be explored. With this, we can possibly reduce the cost of annotations.
Speech-based data augmentation techniques can also be used to increase the amount of data available to us.
\section*{Limitations}
Our work has certain limitations as described in this section. The NSED dataset used in our experiments is small in comparison to some of the publicly available emotion recognition datasets. The number of utterances which do not belong to the \textit{neutral} class is low. Positive classes like \textit{happy} and \textit{excited} constitute less than $2\%$ of our dataset. Negative classes like \textit{anger}, \textit{frustration}, \textit{sad}, \textit{disgust}, and \textit{fear} constitute $37\%$ of our dataset. We also acknowledge that the emotions annotated for each utterance might not be the exact emotion intended by the speaker. The emotion annotations are in accordance with the interpretations of the annotators. The Automatic Speech Recognition (ASR) step is a bottleneck to our pipeline. As all the conversations are code-mixed, code-switched, natural, and often with a lot of noise, the ASR model couldn't generate an accurate transcript sometimes which lead to poor text features and omission of important words for emotion recognition. We acknowledge that the model might have possibly learnt some sensitive customer information. In future, we will include experiments in our study to remove such sensitive information. We also understand that there are state-of-the-art transformer models which can be experimented on. But due to the limited size of our dataset, we couldn't perform those experiments now. In future, as we expand the dataset, we will also include experiments utilising these transformer based models in our study.

\section*{Ethics Statement}

The Natural Speech Emotion Dataset (NSED) dataset used in our experiments was annotated by a team of 4 annotators. Each annotator had to listen to an audio conversation between a customer and a customer-care executive and annotate each speaking turn with emotion, sentiment, valence, arousal, and dominance values. The conversational audio files were provided to us by our industry partner because of which NSED remains a proprietary dataset. Consent was taken from both customers and customer-care executives before recording their conversations. The annotators were paid for the time and effort they spent on the annotation task.



\bibliography{anthology,custom}
\bibliographystyle{acl_industry_natbib}




\end{document}

%% file: sections/intro.tex
\section{Introduction}
Conversational agents which can participate in a dialogue effectively have massive applications across multiple domains. ~\citet{mensio2018rise} discussed three steps of evolution for conversational agents: textual interaction, vocal interaction and embodied interaction. Recently, OpenAI released ChatGPT, a multi-lingual textual conversational model based on the large language model (LLM) GPT 3.5. ChatGPT can “answer follow-up questions, admit its mistakes, challenge incorrect premises, and reject inappropriate
requests” effectively while retaining knowledge from the conversational context as well as the pre-training phase~\citep{bang2023multitask}. ChatGPT has outperformed state-of-the-art LLMs for various tasks in the zero-shot setting.  It was found that, through interactivity, one can improve the performance of ChatGPT by 8\% ROUGE-1 on summarization tasks and 2\% ChrF++ on the machine translation tasks~\citep{bang2023multitask}. With the integration of interactability, ChatGPT has leaped over traditional LLMs with applications across several domains such as law, healthcare, finance and education. 
\\ \\
In many situations, conversation through the speech modality is favorable and convenient as compared to the textual modality. ChatGPT, while a great conversational agent, can only work with the textual modality. A conversational agent which can take speech input and give speech responses that are polite and empathetic, in an end-to-end fashion, is the next phase of evolution for interactive chatbots.
\\ \\
Conversational agents such as ChatGPT need to recognize the emotion of the human interlocutor correctly in order to give responses which are polite and empathetic in nature. Emotion recognition, when done efficiently by chatbots, make the conversations more human-like. Speech emotion recognition is an important sub-task while developing speech-to-speech chatbots.
\\ \\ 
\begin{figure*}
\centering
 \includegraphics[width=\linewidth]{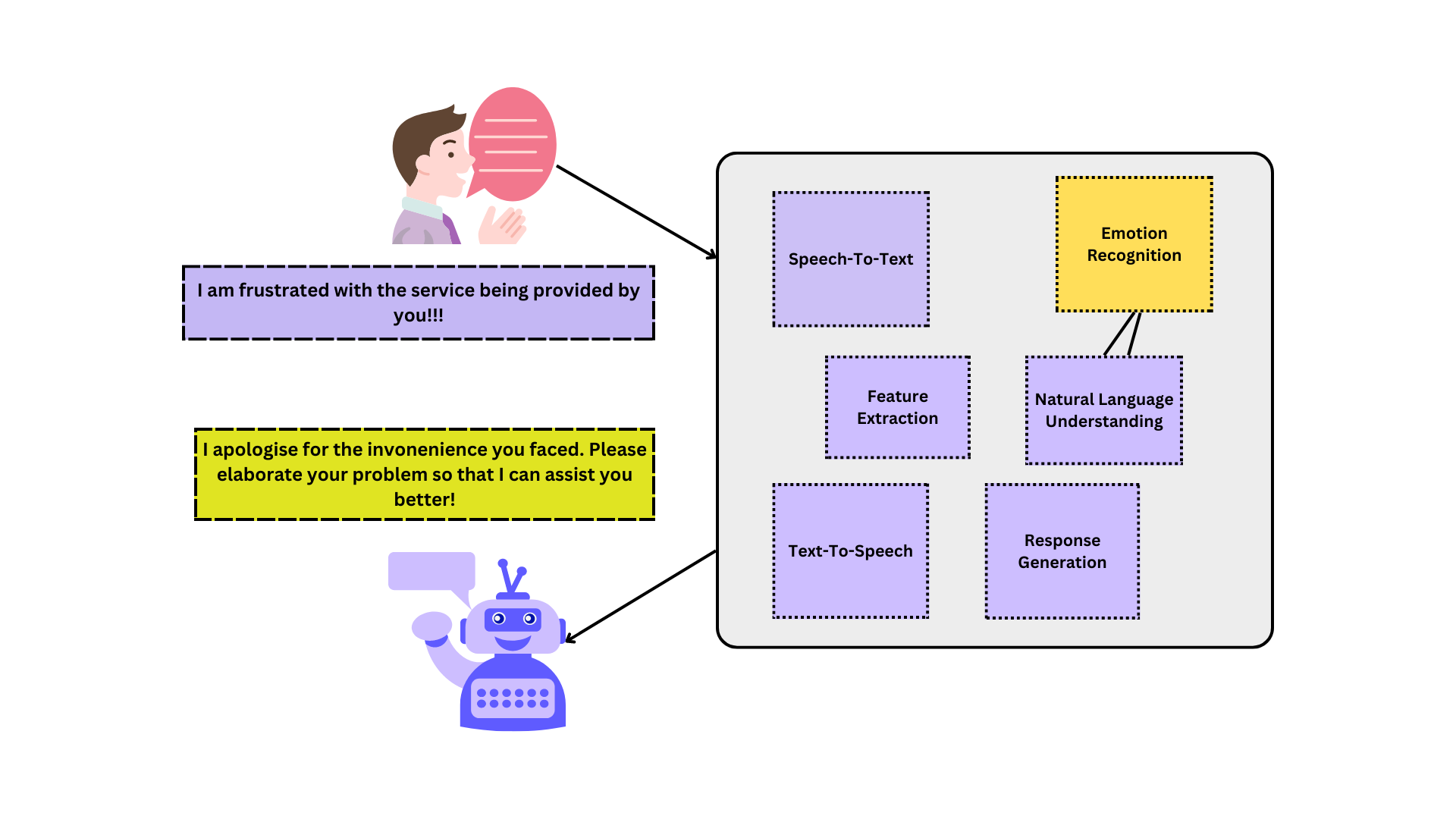}
\caption{Flow-diagram of an ideal conversational agent that generates polite and empathetic responses. Emotion recognition is an essential step in this pipeline.}
\label{fig:serr}
\end{figure*}
Our specific problem statement is to solve Speech Emotion Recognition (SER) where the input is the raw audio of a spoken utterance in a dyadic conversation and the output is its corresponding emotion label, valence, arousal and dominance for a natural code-mixed speech dataset. 
\textbf{Speech Emotion Recognition (SER)} is the task of identifying the emotion of a spoken utterance. Dimensional models plot emotions across the three dimensions of \textit{\textbf{arousal, dominance}} and \textit{\textbf{valence}}. Arousal, valence and dominance signify the intensity, polarity and control exerted by an emotion, respectively. For example, \textit{anger} has high arousal, negative valence and high dominance whereas \textit{fear} has low arousal, negative valence and low dominance. Categorical models define discrete emotion classes such as \textit{anger, happy} and \textit{sad} for various downstream tasks. 
\\
    
Our contributions are:
\begin{enumerate}
    \item A model trained on a natural code-mixed speech emotion dataset, Natural Speech Emotion Dataset (NSED), for the task of Speech Emotion Recognition (SER). NSED has over $5000$ conversational utterances annotated for emotion, sentiment, valence, arousal, and dominance. 
    \item The technique of incorporating word-level VAD values to improve the performance for SER by 2\% for negative emotions, in an industry setting. High accuracy for negative emotion recognition is essential, because customers expressing negative opinions/views need to be pacified with urgency, lest complaints and dissatisfaction snowball and get out of hand. Escalation of negative opinions speedily is crucial for business interests.
\end{enumerate}

\subsection{Motivation}
SER has been an important yet challenging task for researchers. Whenever there is a human-machine interaction in environments where only speech can be propagated, SER becomes a key step for the machine to generate an appropriate response. The task of Emotion Recognition in Conversation (ERC) has many controlling variables such as the context, topic, argumentation logic and speaker/listener personalities, describe the emotional states of the interlocutors. \\ \\
A recent study ~\citep{catania2022conversational} explored the benefits of using an emotion-aware chatbot to help people with alexithymia, a condition which makes it difficult to understand and express emotions. Alexithymia is common in people with neurodevelopmental disorders (NDD). The chatbot provided different utterances to the users and asked them to imitate those utterances by inducing some kind of emotion such as joy or anger. It was found that the interaction with the chatbot became more straightforward as users acquired familiarity: 17 of the 19 participants could perform all emotional activities with progressively decreasing help from the facilitator.  
\\ \\
Most of the SER datasets available today are created by employing professional actors in a clean noise-free environment. In a natural setting, conversations are impromptu, often involving frequent code-mixing and code-switching between multiple languages such as Hindi, English, Marathi, etc. In a customer care setting, it is essential for conversational agents to be polite and empathetic in response to the emotion expressed by the customer. This leads to better overall customer satisfaction and customer retention rates. 
\\ \\
Our \textbf{industry-partner} is a unicorn company in the  Conversational AI sector which empowers over 45000 businesses across the world through their conversational messaging platform. This platform helps businesses engage with customers effectively across commerce, marketing and support with over 9 Billion messages per month. Their mission is to "build the most advanced and innovative platform for conversational engagement with a focus on delivering customer delight". 
\\ \\
We are collaborating with them to work on speech emotion recognition. Through our discussions with them, we explored various ways to approach this problem. They gave us a clear picture of the real-world challenges that are existent in the conversational AI sector. Some of the major challenges are: frequent code-mixing, low-quality recordings and a lack of annotated natural conversational datasets. 
\\ \\
As we will discuss further, the dataset annotated for our experiments, NSED, contains customer care conversations from the escalation department of a customer care service. High accuracy for negative emotion recognition is essential, because customers expressing negative opinions/views need to be pacified with urgency, lest complaints and dissatisfaction snowball and get out of hand. Escalation of negative opinions speedily is crucial for business interests. This tells us that a speech emotion recognition model operating for the escalation department should be very good in detecting negative emotions in conversations. 
\\ \\
An SER model which is capable of capturing contextual information well and is robust to the variations introduced by a natural code-mixed conversation dataset needs to be developed. This model then can be utilised in making speech-to-speech conversational agents more polite and empathetic in an escalation department setting. Figure~\ref{fig:serr} depicts the importance of emotion recognition while developing emotion aware conversational agents.\\



%% file: sections/related.tex
\section{Related work}
Traditionally acoustic speech features have been used along with a statistical machine-learning model for the task of SER ~\citep{schuller2003hidden}. However, selecting the appropriate combination of these low-level features for any given task demands a lot of domain knowledge. Pre-trained deep learning based models trained for other speech processing tasks such as ASR were fine-tuned for SER to get better results ~\citep{lu2020speech}. Recently, self-supervised techniques such as Wav2Vec 2.0~\citep{baevski2020wav2vec} have emerged which learn appropriate speech representations automatically for speech recognition. In~\citet{pepino2021emotion} learned speech representations from Wav2Vec 2.0 are utilized in a downstream model for speech emotion recognition. The proposed model outperformed the state-of-the-art for IEMOCAP~\citep{busso2008iemocap} and RAVDESS~\citep{livingstone2018ryerson} datasets. The study also showed that combining low-level acoustic features with the Wav2Vec 2.0 speech representations resulted in performance gains. In ~\citet{poria2019emotion} it was shown that detecting an emotional shift in conversations is still a bottleneck for SER. In ~\citet{tian2015emotion} non-verbal features were combined with low-level descriptors to improve the performance of emotion recognition in dialogue conversations of the IEMOCAP dataset. In~\citet{vaudable2012negative} the impact of negative emotions on the quality of a call center dialogue was investigated. A study has shown that including dialogue features such as turn number, the topic of discussion ad customer/agent response time can significantly improve the performance of text-based emotion recognition systems~\citep{herzig2016classifying}. In~\citet{han2020ordinal} it was shown that by converting a categorical SER task to an ordinal SER task performance for SER can be improved for customer care calls. ~\citet{deschamps2022investigating} showed that using transformer-based architectures like Wav2vec2-xlsr-53 (for speech) and FlauBERT (for text) increase the performance accuracy by over $20\%$ over baselines. Late fusion of speech and text features also showed performance gains for the task of SER. ~\citet{kulkarni-bhattacharyya-2021-retrofitting} showed that by retrofitting VAD values into word-embeddings one can generate embeddings which are more emotion-aware. A recent study showed that utilising VAD-values and a muti-task framework with emotion recognition as the main task and intensity prediction as the auxiliary task improved performance of emotion recognition on suicide notes~\citep{ghosh2023vad}. Today, using transformer-based architectures like Wav2vec2 and BERT and fusing features of different nature give the best results for SER.   

%% file: sections/modeling.tex
\section{Modeling}
Our model can be mathematically represented using the below argmax equation.
\begin{equation} \label{maths_eqn}
E^* = \argmax\limits_{E} P(E|<F>, <VAD>)
\end{equation}
Here, $E^*$ is the emotion class that maximizes the probability function given a feature set, $<F>$, word-level VAD values of an utterance, $<VAD>$. Our work aims to show that including the feature set $<VAD>$ improves the performance of SER for a natural code-mixed dataset. 

\section{Block Diagram and Architecture}
\begin{figure*}[h]
\centering
\includegraphics[width=\textwidth]{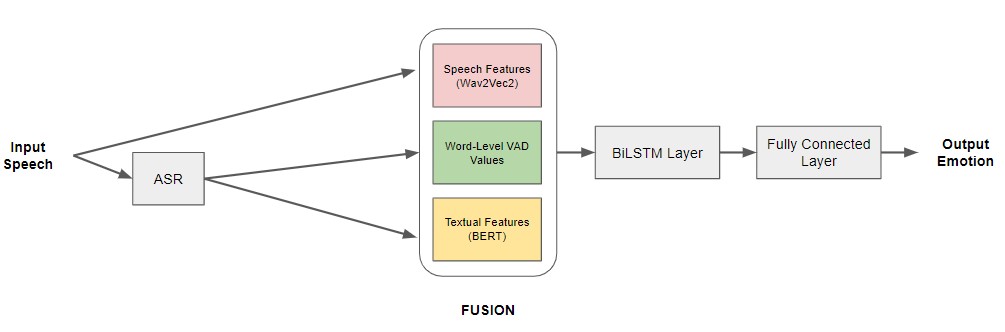}
\caption{Overall architecture of the proposed model. Speech features are extracted from the Wav2Vec2 model. Automatic Speech Recognition (ASR) is used to generate transcripts from the speech input. Word-level valence, arousal, dominance (VAD) values, and textual features are extracted from the ASR transcripts. Fused features are then passed through a BiLSTM layer and a fully-connected layer to finally produce an emotion prediction.}
\label{figure1}
\end{figure*}

In Figure~\ref{figure1}, the overall architecture of the proposed technique is presented. Speech-based features are extracted using the Wav2Vec2 model. Textual features are extracted from the ASR transcripts using the multilingual-BERT model. Word-level valence, arousal, and dominance (VAD) values are extracted from the ASR transcripts using the NRG-VAD lexicon. All these features, once extracted, are fused together and fed into a BiLSTM model. Then a fully-connected layer along with the softmax layer is used to finally generate the predicted emotion.

%% file: sections/experiment.tex
\section{Datasets}
Customer care conversations were recorded and annotated for emotion recognition. The annotation methodology followed is described below.
\subsection{Natural Call Center Speech Emotion Dataset}
Natural Speech Emotion Dataset (NSED) is a code-mixed dyadic customer care conversation dataset created in collaboration with our industry partner. Below are the steps followed to create this dataset. 

 
 
\begin{itemize}
    \item \textbf{Data Recording:} Our industry partner provided us with over 18000 dyadic customer care audio recordings with duration ranging between a few seconds to about an hour and their corresponding machine-generated text transcripts. All the audio recordings were single-channel (mono) with a sampling rate of 8000Hz. The conversations are interactions between a customer and a customer care executive from the complaint escalation team of a car servicing company. Both the speakers, in most of the audio recordings, switch between Hindi and English freely with some occasional use of regional words in languages such as Marathi.
    \item \textbf{Data Processing:} Thirty audio recordings were chosen, each of which was 8-10 minutes long making a total of 4.5 hours long audio recordings. The audacity tool was used to process audio files. Each of these audio recordings was clipped into smaller audio clips corresponding to each \textbf{speaking turn}. A speaking turn is defined as the utterance corresponding to a particular speaker before and after any other speaker speaks. Each of these audio clips were then aligned with their corresponding machine-generated transcripts and were tagged with either "customer" or "executive" depending on who was speaking. The machine-generated transcripts contained many crucial mistakes such as wrongly transcribing the word "escalation" as "cancellation". So, the transcripts were corrected, manually, in order to achieve a better quality of textual data. In some instances, the audio quality drops drastically, making it very difficult to understand the words that are being spoken. In this case, a tag, \textbf{<inaudible>} is used in place of its transcript and further annotations are not performed.
    \item \textbf{Emotion Annotation:} The emotion annotations were performed by a group of annotators with a graduate degree, proficient in both English and Hindi. The annotators worked in pairs to listen and annotate these clips with emotion (neutral, happy, sad, excited, anger, fear, surprised, frustrated, disgust), sentiment (neutral, positive, negative), valence, arousal and dominance (VAD). VAD values were annotated in a scale from 1 to 10 where (5, 5, 5) corresponds to the VAD values of a completely neutral emotion. For VAD, 1 represents the minimum value and 10 represents the maximum value any of the dimensions can have e.g. for valence, 1 represents the most negative and 10 represents the most positive any emotion can get. As we can represent 1000 emotions using the VAD dimensional model and only 9 using the categorical emotion model, not all utterances tagged as "neutral" will have VAD values of (5, 5, 5). For a subset of dataset, consisiting of 1989 utterances, annotated by pair of annotators, the inter-annotator agreement was found to be 0.33 and 0.37 for emotion and sentiment labels respectively by using the cohen-kappa metric of agreement.
    \item \textbf{Dataset Examples:} Two of the examples from the annotated NSED dataset are given below:
    \subitem{\textbf{Example 1:}} The utterance is \textbf{"Do you want someone to get arrested? Haan?"} and its corresponding emotion, sentiment, valence, arousal and dominance are respectively- \textbf{\textit{anger, negative, 2, 8, 9}}.
    \subitem{\textbf{Example 2:}} The utterance is \textbf{"Mai samajhta hun aapko jo bhi problem hui hai. Aage se aapko ye nahi hoga nischint rahiye."} and its corresponding emotion, sentiment, valence, arousal and dominance are respectively- \textbf{\textit{neutral, positive, 6, 5, 5}}.
    
\end{itemize}
\begin{table}
\centering
\begin{tabular}{|c|c|c|}
\hline
\textbf{Emotion} & \textbf{Utterance Count} & \textbf{\%age} \\ \hline
Neutral          & 3510 & 61\%                   \\ \hline
Anger            & 863 & 15\%                     \\ \hline
Frustration     & 748 & 13\%                         \\ \hline
Disgust          & 116 & 2\%                     \\ \hline
Sad              & 403 & 7\%                    \\ \hline
Fear            & 19 & < 1\%                        \\ \hline
Happy            & 57 & < 1\%                    \\ \hline
Surprised            & 13 & < 1\%                    \\ \hline
Excited            & 25 & < 1\%                    \\ \hline
\textbf{Total}   & \textbf{5754} & 100\%                     \\ \hline
\end{tabular}
\caption{\label{table1}Per-emotion distribution of the Natural Speech Emotion Dataset (NSED). The dataset contains \textit{Neutral} utterances in the majority ($61\%$). Negative emotions like \textit{Anger}, \textit{Frustration}, \textit{Disgust}, \textit{Sad}, and \textit{Fear} constitute $37\%$ of the dataset. Positive emotions like Happy, Surprised and Excited constitute $2\%$ of the dataset.}
\end{table}
\begin{figure*}[h]
\centering
\includegraphics[width=\textwidth]{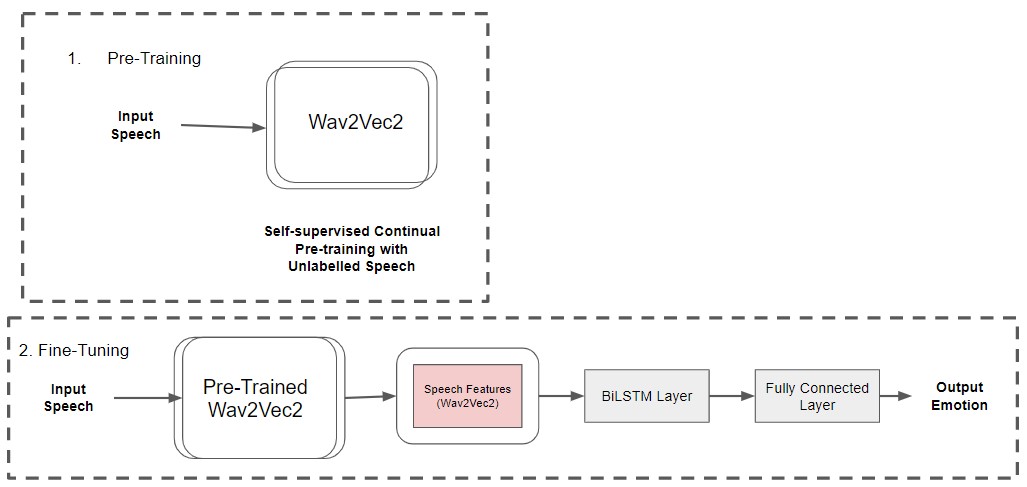}
\caption{Continual pre-training and fine-tuning of the Wav2Vec2 architecture with unlabeled and labeled NSED data, respectively.}
\label{fig:pt}
\end{figure*}

\section{Methodology}

Text features, Wav2vec2 features, and word-level VAD values are extracted and fused together. Indic-Wav2Vec2 is used to extract speech features that constitute a 768-dimensional vector. Whisper-large~\citep{radford2022robust} is used to generate transcripts for each utterance in a conversation. The multi-lingual BERT model is used to generate textual embeddings for each utterance resulting in a 768-dimensional vector. The fused features are then passed through a BiLSTM layer and a fully-connected layer. Finally, a softmax layer is used to predict the corresponding emotion for an utterance. Before extracting the speech features, the wav2vec2 architecture is continually pre-trained as described below.
\begin{table}[]
\centering
\fontsize{10}{12}\selectfont
\begin{tabular}{|p{3cm}|c|c|c|}
\hline
\textbf{Model Name}& \textbf{Neu}& \textbf{Pos}& \textbf{Neg} \\ \hline
Wav2Vec2-XLSR (w/o pre-training)                      & 0.81 & 0.05 &  0.40                            \\ \hline
Wav2Vec2-XLSR (with continual pre-training)                      & 0.89 &  0.05 & 0.53                           \\ \hline
Indic-Wav2Vec2 (w/o pre-training)                      & 0.92 &  0.10 & 0.57                           \\ \hline
\textbf{Indic-Wav2Vec2 (with continual pre-training)}                      & \textbf{0.92} &  \textbf{0.14} & \textbf{0.61}                           \\ \hline
\end{tabular}
\caption{\label{table2} Results for \textbf{Wav2Vec2-xlsr} and \textbf{Indic-Wav2Vec2} with and without continual pre-training with unlabelled audio files from NSED. For the \textbf{Neutral (Neu)} class, precision values are given while for the \textbf{Positive (Pos)} and the \textbf{Negative (Neg)} classes, weighted average precision is given.}
\end{table}
\subsection{Pre-training Wav2Vec2} 
Data annotation is a cost-intensive task that is not feasible to do for the whole unlabeled speech dataset (\textasciitilde$18000$ customer care audio files) provided by our industry partner. Wav2vec2 is a self-supervised speech model which learns speech representations from raw audio signals directly. These speech representations have been shown to be very useful for several speech-processing tasks. Wav2vec2 is pre-trained on 52,0000 hours of Librispeech dataset because of which it has already learned various characteristics of speech present in that dataset. To get even better representations for our dataset we apply a technique called continual pre-training where we continue the pre-training phase with our own unlabeled speech dataset. ~\citet{kessler2022adapter} show that using an adapter-based continual pre-training approach for the wav2vec2 architecture reduces computational cost significantly. We use a similar approach to pre-train the Wav2vec2 architecture using the unlabelled NSED dataset. After pre-training, the Wav2Vec2 architecture is fine-tuned for NSED to evaluate the performance for SER with and without continual pre-training. Table~\ref{table2} shows the precision for the neutral class and the weighted average precision for negative and positive emotions for Wav2Vec2-xlsr and Indic-Wav2Vec2~\citep{javed2021building}. Indic-Wav2Vec2 gives the best performance with continual pre-training. We use this continually pre-trained indic-wav2vec2 model for our experiments. Figure~\ref{fig:pt} shows the pipeline of continual pre-training utilised in our experiments.

\section{Experimental Setup} 
The NSED dataset was split into \textit{train}, \textit{dev}, and \textit{test} sets in the proportions of $80\%$, $10\%$, and $10\%$ respectively. In each experimental run, the dataset was shuffled with a different seed value before feeding it to the model. The NViDia RTXA6000 GPU was used for all the experiments. A single experimental run took approximately 1 hour to complete. Hyper-parameter tuning was performed using the random search technique. The hyper-parameters which gave the best overall performance for the negative emotions were used in the end. The results shown in this paper give the performance of the best experimental run for the negative emotions in terms of weighted-average precision.

%% file: sections/results.tex
\section{Results and Analysis}
Table~\ref{table3} gives the performance of the BiLSTM model using different types of features. 
\subsection{Analysis}
Only by using the Wav2Vec2 (W) features, our model achieves an average precision of $0.61$ over all the negative emotions. This forms the baseline for our experiments. When both the Wav2Vec2 (W) and the textual BERT (T) features are concatenated together, our model achieves a weighted-average precision of $0.64$ over all negative emotions. This shows that textual features have additional emotional information which is absent from only the speech features. When word-level VAD values (VAD), extracted from the NRG-VAD lexicon, are also concatenated along with Wav2Vec2 (W) and textual BERT (T) features, we see an improvement of $2\%$ with a weighted-average precision of $0.66$ over all the negative emotions. This shows that by utilizing word-level VAD values we can improve the performance of our SER model for negative emotions. For the neutral emotion class, all the models achieve a precision over $90\%$. Results for positive emotions are unsatisfactory, with our proposed model giving a weighted-average precision of $0.16$ for all the positive emotions. This can be attributed to the low amount of utterances with positive emotion in NSED. Even though the performance of our model is poor for positive emotions, it performs well for negative emotions, which is ideal, as we are dealing with customer call conversations where the customer is usually unsatisfied with a product or a service.
\begin{table}[]
\fontsize{9}{11}\selectfont
\centering
\begin{tabular}{|c|c|c|c|c|c|c|}
\hline
\textbf{Model Name}& \textbf{Neu}& \textbf{Ang}& \textbf{Sad} & \textbf{Fru}&  \textbf{Neg} & \textbf{Pos} \\ \hline
W (Baseline)                      & 0.92 &  0.74 & 0.63 & 0.69 & 0.61 & 0.14                           \\ \hline

T+W                      & 0.93 &  0.76 &  0.64 & 0.71 & 0.64 & 0.15                           \\ \hline

W+VAD                      & 0.95 & 0.75  & 0.64 & 0.71 & 0.65 & 0.15                            \\ \hline

T+VAD                      & 0.95 & 0.78  & 0.65 & 0.72 & 0.65 & 0.15                            \\ \hline

\textbf{T+W+VAD}                      & \textbf{0.96} & \textbf{0.79} & \textbf{0.67} & \textbf{0.74} & \textbf{0.66} & \textbf{0.16}                             \\ \hline


\end{tabular}
\caption{\label{table3}Results for the proposed model trained using \textbf{Speech Features (S)}, \textbf{Wav2vec2 Features (W)},  \textbf{Textual Features (T)},  and \textbf{VAD values (VAD)}. The last two columns give the \textbf{weighted-average precision} over all the negative emotions and positive emotions. }
\end{table}

\subsection{Challenges}
We faced a number of challenges that one might expect while dealing with a natural code-mixed speech dataset. Some of the challenges are described below:
\begin{itemize}
    \item \textbf{Audio Quality}: Poor quality of the audio recordings made our task even more challenging. Due to network irregularities many recordings' audio quality dropped drastically making it hard for the annotators to annotate properly. The call recordings were created in a single-channel format (mono) which made it difficult to segregate audio clips if two people spoke simultaneously.
    \item \textbf{Transcription Errors}: Our ASR model struggled with the constant code-switching and noisy environments to produce coherent transcriptions for the spoken utterances. These errors then reflected into poor textual embeddings and missing word-level VAD values.
    \item \textbf{Neutral Utterances}: As shown in the table~\ref{table1}, $61\%$ of the utterances in our dataset are neutral in nature. Because of this, our model was more biased in predicting the neutral class than any other emotion classes.
    \item \textbf{Frequent code-mixing and code-switching}: Code-mixing and code-switching make it difficult to extract good features. For the speech input, there doesn't exist a Wav2vec2 model fine-tuned on Hindi+English code-mixed data. The multi-lingual model, Wav2vec2-xlsr-53, fine-tuned for Hindi was used to generate speech representations. The text generated after ASR was transliterated to Hindi. The multilingual BERT-large model was used to generate textual embeddings for the transliterated Hindi text. This transliterated Hindi text was also used to find word-level VAD values. With transliteration as the bottleneck, VAD values for many words weren't found from the NRC-VAD lexicon.
\end{itemize}